\documentclass[letterpaper]{article} 
\usepackage{aaai2027}  

\usepackage[hyphens]{url}  
\usepackage{graphicx} 
\usepackage{natbib}  
\usepackage{caption} 
\usepackage{algorithm}
\usepackage{algorithmic}
\usepackage{amsmath,amssymb}
\usepackage{multirow}
\usepackage{array}
\nocopyright

\usepackage{newfloat}
\usepackage{listings}
\DeclareCaptionStyle{ruled}{labelfont=normalfont,labelsep=colon,strut=off} 
\floatstyle{ruled}
\newfloat{listing}{tb}{lst}{}
\floatname{listing}{Listing}

\usepackage{booktabs}

\title{
Beyond Boundary Frames: Talking-Head Inbetweening via\\
Context-Aware Motion Modeling
}
\author{
Yuchen Deng\textsuperscript{\rm 1,\rm 2},
Hai-Tao Zheng\textsuperscript{\rm 1,\rm 2},
Jie Wang\textsuperscript{\rm 1,\rm 2},
Xiaotian Li\textsuperscript{\rm 1},\\
Feidiao Yang\textsuperscript{\rm 2}\corresponding,
Yuxing Han\textsuperscript{\rm 1}\corresponding
}

\affiliations{
\textsuperscript{\rm 1}Tsinghua Shenzhen International Graduate School, 
Tsinghua University\\
\textsuperscript{\rm 2}Pengcheng Laboratory\\
\{dyc23,jie-wang24\}@mails.tsinghua.edu.cn\\
\{zheng.haitao,yuxinghan\}@sz.tsinghua.edu.cn\\
xt-li23@tsinghua.org.cn, yangfd@pcl.ac.cn
}

\begin{document}

\maketitle

\begin{abstract}
Existing talking-head generation methods primarily target open-ended generation rather than bridging two existing video segments.
In this paper, we study \textbf{talking-head inbetweening}, a practical editing task that aims to generate realistic intermediate frames under fixed endpoint constraints.
Unlike generic video inbetweening, this task requires recovering subtle speech-driven facial dynamics over long temporal gaps, where the boundary frames alone provide insufficient guidance for realistic motion recovery.
To address this problem, we propose \textbf{BBF (Beyond Boundary Frames)}, a unified context-aware framework for talking-head inbetweening. 
BBF consists of three complementary components: Endpoint Anchoring for preserving endpoint consistency, Motion Evolution Modeling for capturing plausible temporal transitions from surrounding visual context, and Speech Dynamics Refinement for injecting fine-grained speech-driven facial dynamics from speech audio.
A progressive optimization strategy further balances structural consistency and motion refinement during denoising.
Extensive experiments on the talking-head benchmarks HDTF and Hallo3 demonstrate that BBF consistently achieves state-of-the-art performance. In particular, BBF surpasses the strongest baseline on Hallo3 by 23.3\% in FID and 36.5\% in FVD. 
Moreover, BBF demonstrates strong generalization on generic video inbetweening benchmarks.

\end{abstract}


\section{Introduction}
\label{sec:Introduction}

\begin{figure}[t]
    \centering
    \includegraphics[width=\linewidth]{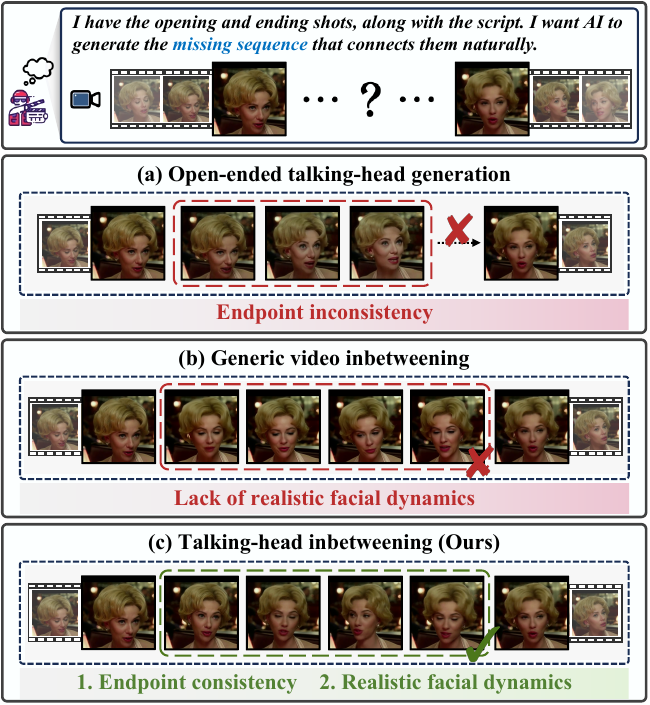}
    \vspace{-5mm}
    \caption{Comparison of talking-head generation, generic video inbetweening, and talking-head inbetweening.}
    \label{fig:Introduction}
    \vspace{-4mm}
\end{figure}

Talking-head generation has achieved remarkable progress with the emergence of large-scale video diffusion models. Recent methods can synthesize highly photorealistic talking videos with realistic lip synchronization, expressive facial motions, and strong identity preservation~\cite{wei2024aniportrait,tu2025stableavatar,tu2026flashportrait}. These advances have enabled a wide range of applications~\cite{team2025kling}, including digital avatars, virtual presenters, and other human-centric content creation scenarios.

Despite their impressive performance, existing methods are primarily designed for open-ended video generation rather than practical video editing. 
They typically generate videos by progressively synthesizing future frames from a single reference image conditioned on audio, without leveraging future visual observations. 
In many editing scenarios, however, users need to \emph{bridge} two existing talking-head video clips instead of generating an entire video from scratch. As illustrated in Fig.~\ref{fig:Introduction}, this requires generating intermediate frames that establish a seamless transition between the two clips while remaining consistent with both boundary frames.
Such a capability is essential for practical video editing tasks~\cite{briedis2021neural}, such as film post-production and digital avatar editing, where facial expression correction and local content modification are frequently required.

Motivated by this practical need, we study \textbf{talking-head inbetweening}, a largely unexplored video editing task that aims to generate intermediate frames that seamlessly bridge two existing talking-head clips while satisfying fixed endpoint constraints.

Existing video interpolation~\cite{dong2023video,zhang2024vfimamba,zhang2025eden} and inbetweening~\cite{wan2025wan,choi2026anchoring,jeon2026motion} methods primarily infer intermediate motion from the two boundary frames. 
While this strategy is often effective for \emph{generic} scenes, where the primary objective is to recover object trajectories and geometric motion, it becomes insufficient for talking-head videos. 
Unlike generic scenes, talking-head videos require modeling fine-grained facial dynamics, such as lip articulation and facial expressions. 
Although the boundary frames constrain the start and end facial states, they provide only sparse cues about the temporal evolution of these dynamics. 
As a result, relying solely on the boundary frames leaves the temporal evolution of facial dynamics highly under-constrained, often leading to unrealistic intermediate motions (Fig.~\ref{fig:Introduction}). 

To disambiguate the under-constrained facial motion between the two boundary frames, additional contextual information is required. 
Specifically, adjacent video clips reveal the motion evolution surrounding the missing segment, speech provides fine-grained articulatory cues, and text conveys high-level semantic intent.
These cues impose complementary constraints on motion generation.
However, generating intermediate frames that preserve endpoint consistency while effectively exploiting these contextual cues to recover natural facial motion remains challenging. 

To address these challenges, we propose \textbf{Beyond Boundary Frames (BBF)}, a unified framework for talking-head inbetweening that jointly models complementary constraints from multimodal contextual cues.
Specifically, \textbf{Endpoint Anchoring} continuously injects the two boundary frames as temporal anchors throughout the denoising process, preserving structural coherence and appearance consistency under fixed endpoint constraints.
\textbf{Motion Evolution Modeling} explicitly models motion evolution from the adjacent video clips and dynamically modulates motion priors according to the temporal distance of each intermediate frame to the two endpoints, enabling coherent long-range facial motion transitions.
\textbf{Speech Dynamics Refinement} progressively aligns speech dynamics with the denoising trajectory, allowing speech guidance to focus on increasingly fine-grained articulatory motions as generation proceeds.
Finally, we introduce a progressive optimization strategy that gradually shifts the learning emphasis from structural consistency to temporal dynamics, enabling different contextual cues to contribute in a balanced and complementary manner.

Extensive experiments on the talking-head benchmarks HDTF and Hallo3 demonstrate that BBF consistently achieves state-of-the-art performance in both visual fidelity and temporal consistency. In particular, on Hallo3, BBF outperforms the strongest baseline by 23.3\% in FID and 36.5\% in FVD. Moreover, despite being designed for talking-head inbetweening, BBF generalizes effectively to \emph{generic} video interpolation benchmarks, achieving competitive or superior performance on DAVIS and HDTF.

We summarize the contributions of this paper as follows.
\begin{itemize}

\item We study talking-head inbetweening, a practical yet under-explored setting that recovers realistic intermediate frames by leveraging multimodal contextual cues beyond the boundary frames.

\item We propose BBF, a unified framework that jointly models complementary multimodal constraints for realistic and temporally coherent intermediate frame generation.

\item We propose Endpoint Anchoring, Motion Evolution Modeling, and Speech Dynamics Refinement, together with a progressive optimization strategy that balances structural consistency and fine-grained temporal dynamics throughout denoising.

\item BBF achieves state-of-the-art performance on talking-head benchmarks while generalizing effectively to generic video interpolation benchmarks.

\end{itemize}

\begin{figure*}[t]
  \centering
  \includegraphics[width=\textwidth,keepaspectratio]{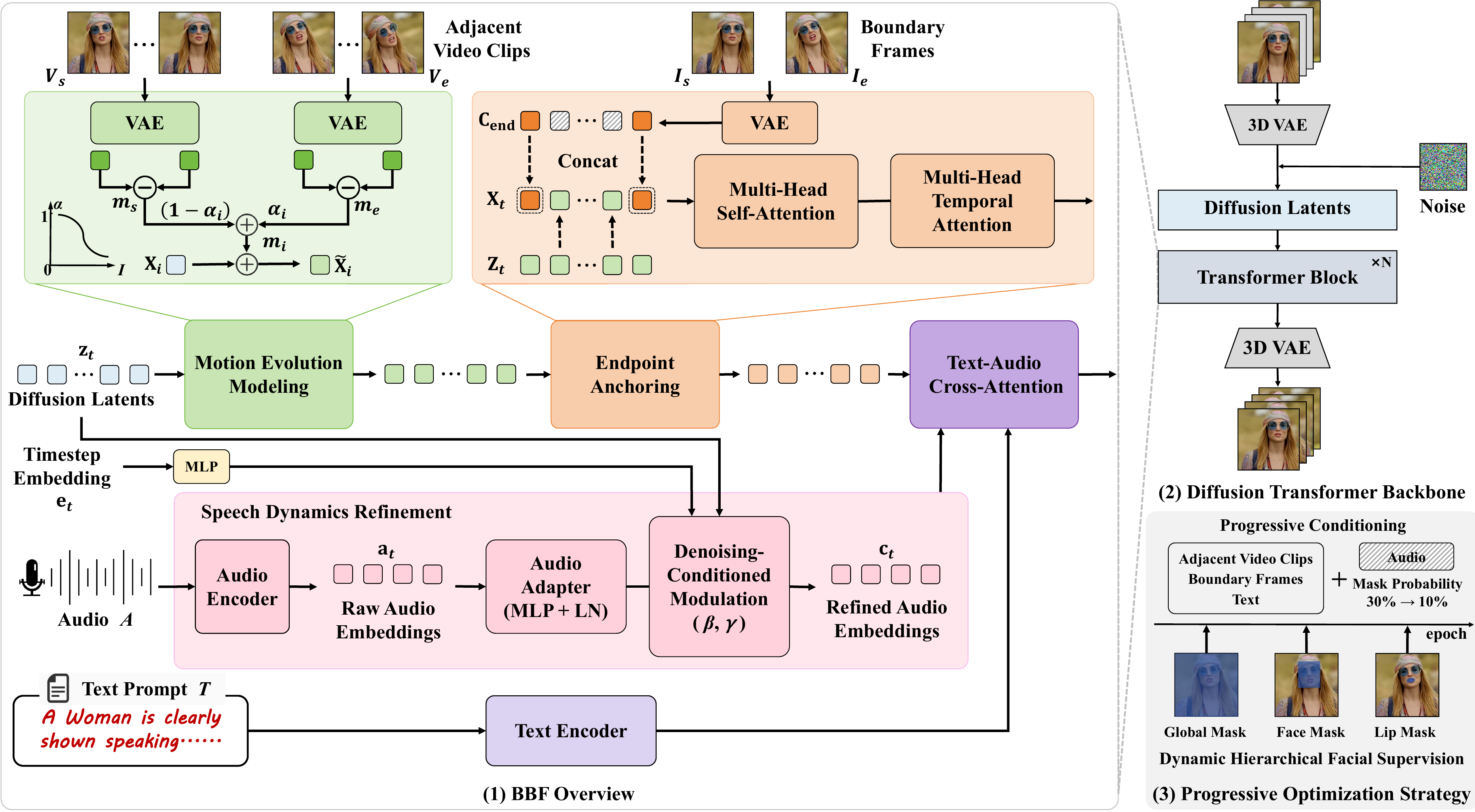}
  \caption{
Overview of the proposed BBF framework.
Given adjacent video clips, boundary frames, speech audio, and an optional text prompt, 
BBF performs Motion Evolution Modeling to capture coarse motion priors from surrounding video clips, Endpoint Anchoring to maintain boundary consistency throughout generation, and Speech Dynamics Refinement to enhance speech-driven facial dynamics. A progressive optimization strategy further improves robustness and generation quality.
  }
  \label{fig:framework}
  \vspace{-4pt}
  \vspace{-2mm}
\end{figure*}

\section{Related work}
\label{sec:Related work}

\paragraph{Talking-Head Generation.}
Audio-driven talking-head generation aims to animate a reference portrait from speech while preserving identity and lip synchronization. Early methods rely on explicit motion representations such as facial landmarks, motion fields, or neural renderers~\cite{zhou2021pose,wang2021one,zhang2023sadtalker}, whereas recent diffusion-based approaches achieve substantial improvements in visual quality and motion realism by modeling rich audio-visual correlations~\cite{wei2024aniportrait,chen2025echomimic,gan2025omniavatar,tu2025stableavatar,deng2026fluentavatarflickerfreetalkingheadanimation,cui2025hallo3,huang2025live,feng2025streamdiffusionv2}. 
These methods formulate talking-head synthesis as open-ended generation from a reference image and speech, without future visual observations. 
In contrast, talking-head inbetweening requires generating the missing frames between two fixed endpoints. This setting introduces fundamentally different constraints, as the generated frames must simultaneously satisfy both boundary conditions while recovering plausible speech-driven facial dynamics. Existing talking-head generation methods are therefore not directly applicable to this task.

\paragraph{Video Inbetweening.}
Generative video inbetweening synthesizes plausible transitions between distant keyframes. Early methods mainly rely on text prompts or reference images to guide generation~\cite{xing2024dynamicrafter,feng2024explorative,wan2025wan}, while recent approaches introduce explicit motion controls such as trajectories or structural constraints to improve controllability~\cite{wang2024framer,guo2025controllable,tanveer2025multicoin,choi2026anchoring,jeon2026motion,zhu2025generative,raagl2026adaptive}. 
However, these methods are primarily designed for generic video transitions and rely on explicit motion control to guide intermediate-frame generation. 
Different from them, our method addresses the intermediate dynamics are governed by multimodal contextual cues, such as speech, facial expressions, and semantic intent. 

\paragraph{Video Interpolation.}
Video frame interpolation (VFI) traditionally estimates intermediate frames between temporally adjacent inputs using convolutional networks, optical flow, or dynamic kernels~\cite{dong2023video,choi2020channel,kalluri2023flavr,niklaus2020softmax,zhang2023extracting,cheng2021multiple,lu2022video}. 
More recently, diffusion-based methods further improve interpolation quality by modeling temporal denoising processes in latent space~\cite{ho2020denoising,rombach2022high,blattmann2023stable,feng2024explorative,huang2022real,li2023amt,zhang2024vfimamba,zhang2025eden}. 
Compared with conventional VFI, which focuses on short-range interpolation between adjacent frames, talking-head inbetweening addresses long-range transitions, where the intermediate facial dynamics are substantially more ambiguous due to the larger temporal gap between the boundary frames.

\section{Method}
\label{sec:Method}

\subsection{Problem Formulation}

Talking-head inbetweening aims to synthesize a sequence of intermediate frames that naturally bridge two existing talking-head video clips. Unlike conventional video inbetweening, the generated sequence should not only remain consistent with the start and end frames, but also recover realistic facial motions according to the multimodal context.

Formally, given the start and end boundary frames $I_s$ and $I_e$, together with a set of multimodal contextual cues $\mathcal{C}_{ctx}=\{V_s,V_e,A,T\}$,
where $V_s$ and $V_e$ denote the adjacent video clips, $A$ denotes the corresponding speech audio, and $T$ denotes a text description, our goal is to generate an intermediate video
$\hat V_m=\{I_1,\cdots,I_K\}$.
The boundary frames $I_s$ and $I_e$ provide deterministic endpoint constraints, whereas the contextual cues $\mathcal{C}_{ctx}$ provide complementary motion and semantic priors to resolve the ambiguity of intermediate facial motion. Specifically, $\{V_s,V_e\}$ provide local motion evolution cues around the two endpoints, audio $A$ provides fine-grained temporal cues for speech-driven facial dynamics, and text $T$ provides high-level semantic guidance for facial behaviors and scene intent. Each modality in $\mathcal{C}_{ctx}$ is \emph{optional}, which flexibly adapts to different application scenarios by exploiting available contextual cues.

\subsection{Overview}

To address the challenges of jointly satisfying endpoint constraints and multimodal contextual guidance, we propose \textbf{Beyond Boundary Frames (BBF)}, a unified context-aware diffusion framework for talking-head inbetweening.

As illustrated in Fig.~\ref{fig:framework}, BBF is built upon a pretrained video diffusion model with a DiT backbone operating on 3D-VAE latent tokens. The boundary frames are embedded as temporal anchor tokens, while multimodal contextual cues are injected through decoupled cross-attention branches, allowing visual, speech, and semantic information to guide denoising in a complementary manner.

Specifically, BBF consists of three complementary modules. \textbf{Endpoint Anchoring} continuously propagates endpoint constraints throughout denoising. \textbf{Motion Evolution Modeling} models motion evolution from adjacent video clips to provide coarse transition priors. \textbf{Speech Dynamics Refinement} progressively refines fine-grained speech-driven facial dynamics. Finally, a progressive optimization strategy gradually shifts learning from structural consistency to realistic temporal dynamics.

\subsection{Endpoint Anchoring}
Preserving consistency with both boundary frames is fundamental to talking-head inbetweening. BBF encodes boundary frames into persistent endpoint anchor tokens and embeds them into the latent sequence, allowing intermediate representations to continuously access boundary information throughout denoising.

Specifically, let
$\mathbf I_{\mathrm{end}}=[I_s,\mathbf 0,\ldots,\mathbf 0,I_e]$
denote the endpoint-only sequence, and let
$\mathbf M_{\mathrm{end}}$ be its binary mask, whose nonzero
entries occur only at the first and last temporal positions.
At diffusion timestep $t$, given the latent representation
$\mathbf Z_t$, we concatenate the endpoint condition with
$\mathbf Z_t$ before 3D patch embedding:
\begin{align}
\mathbf C_{\mathrm{end}}
&=
[\mathcal R(\mathbf M_{\mathrm{end}})
 \Vert
 \mathcal E(\mathbf I_{\mathrm{end}})],
\label{eq:endpoint_condition_c}
\\
\mathbf X_t
&=
\mathcal P_{3D}
([\mathbf Z_t\Vert\mathbf C_{\mathrm{end}}]).
\label{eq:endpoint_condition_x}
\end{align}
where $\mathcal E$ denotes the 3D-VAE encoder,
$\mathcal R$ aligns the endpoint mask with the latent resolution,
$\Vert$ denotes channel-wise concatenation, and $\mathcal P_{3D}$ denotes 3D patch embedding, yielding the endpoint-augmented DiT input tokens $\mathbf X_t$.

Moreover, we insert an additional temporal attention layer after every self-attention block to explicitly propagate information from the endpoint anchors along the temporal dimension. This design enables the boundary constraints to be progressively propagated to intermediate latent tokens, encouraging the synthesized frames to remain consistent with both endpoints throughout generation.

\subsection{Motion Evolution Modeling}

While the boundary frames specify only the start and end facial states, adjacent video clips reveal how facial motion evolves immediately before and after the missing segment. BBF therefore models these motion evolution cues to provide coarse transition guidance for intermediate synthesis.

We sparsely sample frames from the adjacent video clips, encode them into VAE latents, and compute temporal differences between neighboring samples to estimate motion evolution priors from the preceding and succeeding clips. The resulting motion priors are projected into the DiT hidden space as
\vspace{-1mm}
\begin{equation}
\mathbf m_s=W_p\bar{\mathbf m}_s,\qquad
\mathbf m_e=W_p\bar{\mathbf m}_e,
\end{equation}
where $\bar{\mathbf m}_s$ and $\bar{\mathbf m}_e$ denote the motion evolution priors estimated from the preceding and succeeding clips, respectively, and $W_p$ is a learnable linear projection.

To generate frame-aware motion guidance, the two endpoint priors are interpolated according to the normalized temporal position,
\begin{equation}
\mathbf m_i
=
(1-\alpha_i)\mathbf m_s
+
\alpha_i\mathbf m_e,
\end{equation}
where $\alpha_i$ denotes the relative temporal position of the $i$-th latent frame.
The resulting motion prior is broadcast to all spatial tokens within the same frame and injected through a residual connection,
\begin{equation}
\tilde{\mathbf X}_i
=
\mathbf X_i
+
W_m\mathbf m_i,
\end{equation}
where $\mathbf X_i$ denotes the latent tokens of the $i$-th frame and $W_m$ is a learnable projection matrix. Consequently, the modeled motion evolution provides coarse transition guidance while preserving sufficient flexibility for subsequent speech dynamics refinement.

\subsection{Speech Dynamics Refinement}

While the motion evolution module provides coarse transition guidance, realistic talking-head synthesis further requires fine-grained speech-driven facial dynamics. Directly injecting audio features, however, often introduces cross-modal inconsistency because speech representations are not naturally aligned with the denoising latent space.

To address this issue, BBF introduces an Audio Context Adapter to transform raw speech features into denoising-aware visual representations.
Given Wav2Vec embeddings, the adapter first projects speech features into the latent embedding space,
\vspace{-2mm}
\begin{equation}
\mathbf h_t
=
\mathrm{LN}
(
\mathrm{MLP}
(
\mathbf a_t
)
),
\vspace{-1mm}
\end{equation}
followed by denoising-conditioned modulation,
\vspace{-1mm}
\begin{equation}
\mathbf c_t
=
\gamma(\mathbf e_t,\mathbf z_t)
\odot
\mathbf h_t
+
\beta(\mathbf e_t,\mathbf z_t),
\vspace{-1mm}
\end{equation}
where $\mathbf a_t$ denotes the raw audio embedding, $\mathbf h_t$ is the projected audio feature, $\mathbf z_t$ denotes the current latent state, $\mathbf e_t$ is the diffusion timestep embedding, and $\mathbf c_t$ denotes the conditioned audio feature. 
The conditioned audio features are then aggregated as the sequence $\mathbf C$. 

To complement speech articulation with high-level semantic guidance, an optional text prompt is encoded by a frozen text encoder into text embeddings $\mathbf T$. The conditioned audio sequence $\mathbf C$ and text embeddings $\mathbf T$ are then injected through independent cross-attention branches,
\begin{equation}
\widehat{\mathbf X}_{t}
=
\widetilde{\mathbf X}_{t}
+
\operatorname{CA}
\left(
\widetilde{\mathbf X}_{t},
\mathbf C_t,
\mathbf T
\right).
\label{eq:text_audio_cross_attention}
\end{equation}
where $\operatorname{CA}(\cdot)$ denotes the Text-Audio Cross-Attention module. Consequently, the diffusion latents adaptively exploit complementary semantic and speech cues, enabling semantic descriptions and speech dynamics to jointly refine the generated facial motions.

\begin{table*}[t]
\centering
\begingroup
\setlength{\tabcolsep}{3.2pt}
\renewcommand{\arraystretch}{1.0}

\resizebox{\textwidth}{!}{%
\begin{tabular}{@{}l cccccc cccccc c@{}}
\toprule
\multirow{2}{*}{\textbf{Model}}
& \multicolumn{6}{c}{\textbf{HDTF}}
& \multicolumn{6}{c}{\textbf{Hallo3}}
& \multirow{2}{*}{\textbf{Lat.(s)$\downarrow$}} \\
\cmidrule(lr){2-7}
\cmidrule(lr){8-13}

~
& FID$\downarrow$
& FVD$\downarrow$
& LPIPS$\downarrow$
& PSNR$\uparrow$
& SSIM$\uparrow$
& Sync-D$\downarrow$
& FID$\downarrow$
& FVD$\downarrow$
& LPIPS$\downarrow$
& PSNR$\uparrow$
& SSIM$\uparrow$
& Sync-D$\downarrow$
& \\

\midrule

SadTalker~\cite{zhang2023sadtalker}
& 81.02 & 695.21 & 0.56 & 12.28 & 0.44 & 12.66
& 44.07 & 898.48 & 0.33 & 14.83 & 0.67 & 14.62
& 78.92 \\

AniPortrait (Wei et al.~\citeyear{wei2024aniportrait})
& 46.31 & 480.04 & 0.31 & 16.49 & 0.61 & 13.50
& 91.84 & 703.73 & 0.66 & 6.83 & 0.24 & 14.70
& 575.53 \\

EchoMimic~\cite{chen2025echomimic}
& 53.27 & 484.89 & 0.43 & 14.58 & 0.59 & 12.22
& 115.57 & 1263.43 & 0.69 & 9.19 & 0.36 & 14.30
& 512.71 \\

Sonic~\cite{ji2025sonic}
& 45.02 & 366.21 & 0.26 & 18.05 & 0.64 & 12.70
& 37.57 & 375.92 & 0.32 & 15.32 & 0.64 & \textbf{12.88}
& 190.42 \\

OmniAvatar~\cite{gan2025omniavatar}
& 42.63 & 388.33 & 0.27 & 18.32 & \underline{0.69} & 13.33
& 64.01 & 978.57 & 0.43 & 12.33 & 0.58 & 14.62
& 392.30 \\

StableAvatar~\cite{tu2025stableavatar}
& \underline{31.61} & 318.12 & \textbf{0.21} & \underline{19.68} & 0.68 & 12.99
& 44.85 & \underline{243.53} & \underline{0.28} & \underline{16.34} & 0.69 & 14.19
& 187.00 \\

FT-Wan2.1
& 39.89 & \underline{275.12} & 0.24 & 19.39 & 0.66 & \textbf{11.63}
& \underline{32.48} & 548.18 & 0.27 & 16.68 & \underline{0.70} & \underline{12.91}
& 193.40 \\

\midrule

BBF (Ours)
& \textbf{28.81} & \textbf{244.02} & \textbf{0.21} & \textbf{19.75} & \textbf{0.71} & \underline{12.12}
& \textbf{28.83} & \textbf{154.62} & \textbf{0.22} & \textbf{18.82} & \textbf{0.74} & 13.65
& 221.88 \\

\bottomrule
\end{tabular}%
}
\endgroup

\vspace{-2mm}
\caption{Comparison with talking-head generation models.
BBF and FT-Wan2.1 are conditioned on both boundary frames and audio,
whereas the other methods take only the start frame and audio as input
due to their architectural limitations.
Bold and underlined numbers indicate the best and second-best results,
respectively.}
\label{tab:hdtf-hallo3}
\end{table*}

\begin{table*}[t]
\centering
\resizebox{\textwidth}{!}{%
\begin{tabular}{l ccccc ccccc c}
  \toprule
  \multirow{2}{*}{\textbf{Model}} &
    \multicolumn{5}{c}{\textbf{HDTF}} &
    \multicolumn{5}{c}{\textbf{DAVIS}} &
    \multirow{2}{*}{\textbf{Lat.(s)$\downarrow$}} \\
  \cmidrule(lr){2-6}\cmidrule(lr){7-11}
  & FID$\downarrow$ & FVD$\downarrow$ & LPIPS$\downarrow$ & PSNR$\uparrow$ & SSIM$\uparrow$
  & FID$\downarrow$ & FVD$\downarrow$ & LPIPS$\downarrow$ & PSNR$\uparrow$ & SSIM$\uparrow$ & \\
  \midrule
  TRF~\cite{feng2024explorative}           
                & 13.82 & 450.10 & 0.15 & 21.82 & \textbf{0.78} & 177.08 & 1528.04 & 0.41 & 11.45 & 0.43 & 250.47 \\
  Dynamicrafter~\cite{xing2024dynamicrafter}    
                & 13.49 & \underline{181.53} & 0.15 & 22.48 & \textbf{0.78} & 440.53 & 6665.01 & 0.81 & 10.46 & 0.25 & 352.16 \\
  VACE~\cite{jiang2025vace}          
                & 13.62 & 208.41 & 0.20 & 19.64 & 0.76 & 152.93 & 936.80 & 0.32 & 13.98 & \textbf{0.54} & 178.33 \\
  Framer~\cite{wang2024framer}        
                & 14.95 & 392.22 & 0.21 & 20.21 & 0.76 & 175.92 & 1056.88 & 0.36 & 13.44 & 0.48 & 137.11 \\
  KAB~\cite{choi2026anchoring}           
                & 87.15 & 593.69 & 0.26 & 18.20 & 0.64 & \textbf{141.51} & \underline{869.46} & \underline{0.29} & 13.67 & 0.48 & 1019.58\\
Wan2.1~\cite{wan2025wan}        
                & 11.69 & 338.37 & 0.18 & 20.16 & 0.75 & 181.99 & 2051.07 & 0.42 & 12.53 & 0.47 & 189.51 \\
FT-Wan2.1     
                & \textbf{10.47} & 284.23 & \textbf{0.14} & \underline{22.81} & 0.77 & 158.88 & 1514.49 & \textbf{0.28} & \underline{15.09} & \textbf{0.54} & 192.31 \\
  \midrule
  BBF (Ours)
                & \underline{11.67} & \textbf{174.69} & \textbf{0.14} & \textbf{23.04} & \textbf{0.78} & \underline{147.10} & \textbf{713.01} & 0.30 & \textbf{15.38} & \textbf{0.54} & 201.71 \\
  \bottomrule
\end{tabular}
}
\vspace{-2mm}
\caption{Comparison with video inbetweening models on the talking-head dataset HDTF and the generic-scene dataset DAVIS. 
}
\label{tab:davis_hdtf_results_with_latency}
\vspace{-2mm}
\end{table*}

\subsection{Progressive Optimization}

BBF adopts a progressive optimization strategy that gradually shifts learning from structural consistency to speech-driven facial dynamics, enabling different contextual cues to cooperate throughout training.

\paragraph{Progressive Conditioning.}

Since endpoint and motion cues mainly establish the coarse facial structure, while speech cues provide fine-grained articulation details, we adopt a progressive conditioning strategy during training. Specifically, audio features are randomly masked with a probability of 0.3 in the early stage, encouraging the model to first learn structurally consistent transitions under endpoint and motion guidance. As training proceeds, the masking probability is gradually reduced to 0.1, allowing speech cues to progressively refine lip articulation and subtle facial dynamics while preserving the established motion structure.

\paragraph{Dynamic Hierarchical Facial Supervision.}

Standard reconstruction objectives weight reconstruction errors uniformly across spatial locations, despite facial regions being considerably more important than the background in talking-head synthesis. We therefore introduce a region-aware reconstruction objective with dynamically scheduled supervision weights, allowing the training focus to progressively shift from global facial structures to fine-grained lip articulation.

Specifically, the spatial weight map is defined as
\begin{equation}
\mathbf W
=
1+\lambda_{\mathrm f}(s)\mathbf M_{\mathrm{face}}
+\lambda_{\mathrm l}(s)\mathbf M_{\mathrm{lip}},
\end{equation}
where $\lambda_{\mathrm f}(s)$ and $\lambda_{\mathrm l}(s)$ are training-dependent coefficients controlling the relative importance of the face and lip regions, respectively, and $s$ denotes the current training step.
The weighted reconstruction loss is formulated as
\begin{equation}
\mathcal{L}_{\mathrm{rec}}
=
\mathbb E
\left[
\|
(\mathbf Z_{\mathrm{gt}}
-
\mathbf Z_\theta)
\odot
\mathbf W
\|_2^2
\right].
\end{equation}
where $\mathbf Z_{\mathrm{gt}}$ and $\mathbf Z_\theta$ denote the ground-truth and predicted latent representations, respectively, $\mathbf W$ is the spatial weighting map, and $\odot$ denotes element-wise multiplication.

During the early stage of training, larger weights are assigned to the facial region to encourage globally consistent facial structures. As training progresses, the supervision gradually shifts toward the lip region, enabling finer speech articulation and more realistic speech-driven facial dynamics.

\section{Experiments}
\label{sec:Experiments}

\subsection{Experimental Setting}
\paragraph{Datasets.} We train our model on Celebv-HQ \cite{zhu2022celebv}, which contains 35,666 high-quality celebrity face video clips with a minimum resolution of 512$\times$512. 

To evaluate talking-head inbetweening, we use HDTF \cite{zhang2021flow} and Hallo3 \cite{cui2025hallo3}. HDTF is a high-resolution, in-the-wild talking-head dataset consisting of hundreds of frontal videos and is widely adopted for audio-driven talking-head generation. Hallo3 contains over 70 hours of talking videos, including approximately 50 hours of dynamic talking-head editing footage, providing diverse real-world editing scenarios.

To evaluate the generalization ability of our method beyond talking-head videos, we further conduct experiments on DAVIS \cite{pont20172017}, a widely used benchmark for \emph{generic} video inbetweening with large object motions.

\paragraph{Implementation Details.} We initialize DiT from Wan2.1-I2V 1.3B~\cite{liu2025phantom} and fine-tune BBF on 4$\times$A100-80G GPUs (batch size 8 per GPU). Inputs are $512\times512$ videos with 81 frames, without extra augmentation beyond standard normalization. We use AdamW ($\beta_1=0.9$, $\beta_2=0.999$, weight decay $3\times10^{-2}$) with learning rate $2\times10^{-5}$ and a constant-with-warmup schedule (100 warmup steps), and train for 2000 steps. VAE, CLIP, T5, and Wav2Vec \cite{baevski2020wav2vec} are frozen.
Only DiT attention-related parameters and the audio adapter are updated. For diffusion training, we use FlowMatch Euler with 1000 diffusion timesteps and uniform timestep sampling. For inference, we use 50 denoising steps with GS/PCFG/ACFG = 6.0/5.0/5.0.

\paragraph{Evaluation Metrics.}
We evaluate image and video generation quality using FID \cite{heusel2017gans} and FVD \cite{unterthiner2018towards}, respectively. We further report LPIPS \cite{zhang2018unreasonable}, PSNR \cite{hore2010image}, and SSIM \cite{wang2004image} to measure perceptual similarity, reconstruction fidelity, and structural consistency, respectively. For talking-head generation, we additionally use Sync-D \cite{chung2016out} to evaluate audio-lip synchronization. Finally, we report the inference latency (Lat.), defined as the end-to-end time required to generate an 81-frame video on a single NVIDIA A100 GPU.

\subsection{Comparison with State-of-the-Art Methods}

As there are few existing methods specifically designed for talking-head inbetweening, we compare our method with state-of-the-art approaches from two closely related settings: talking-head generation and generic video inbetweening. Comparisons with additional related methods, including video interpolation models, are presented in the Appendix.

\begin{figure}[t]
    \centering
    \includegraphics[width=\columnwidth]
    {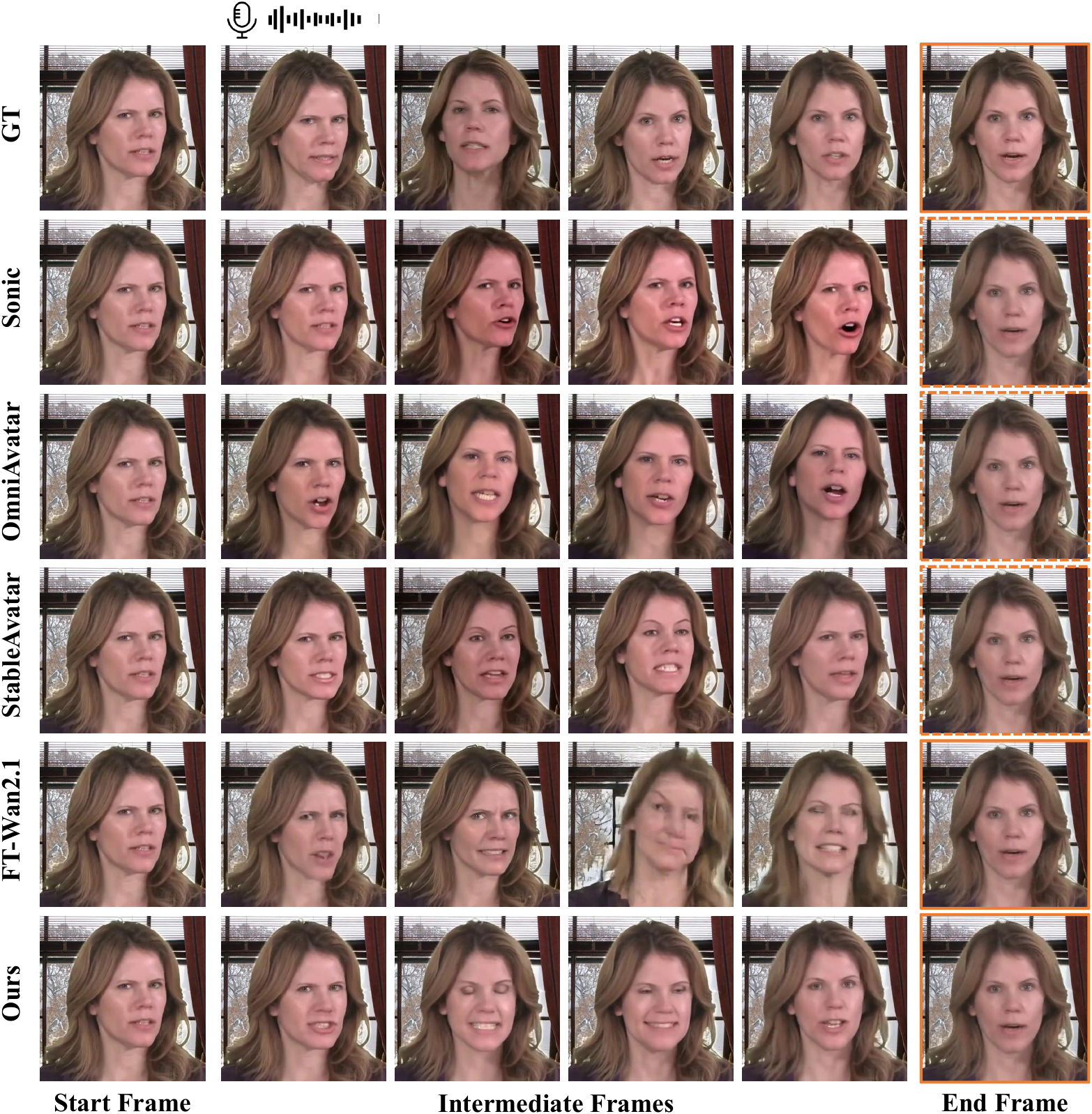}
    \vspace{-6mm}
    \caption{
Qualitative comparison with talking-head generation models. The start frame is a right-profile view, while the end frame is a frontal view. Models that do not support end-frame input (dashed boxes) produce a final intermediate frame that is clearly inconsistent with the target end frame.}
    \label{fig:Qualitative evaluation result}
    \vspace{-4mm}
\end{figure}

\begin{figure}[t]
    \centering
    \includegraphics[width=\columnwidth]
    {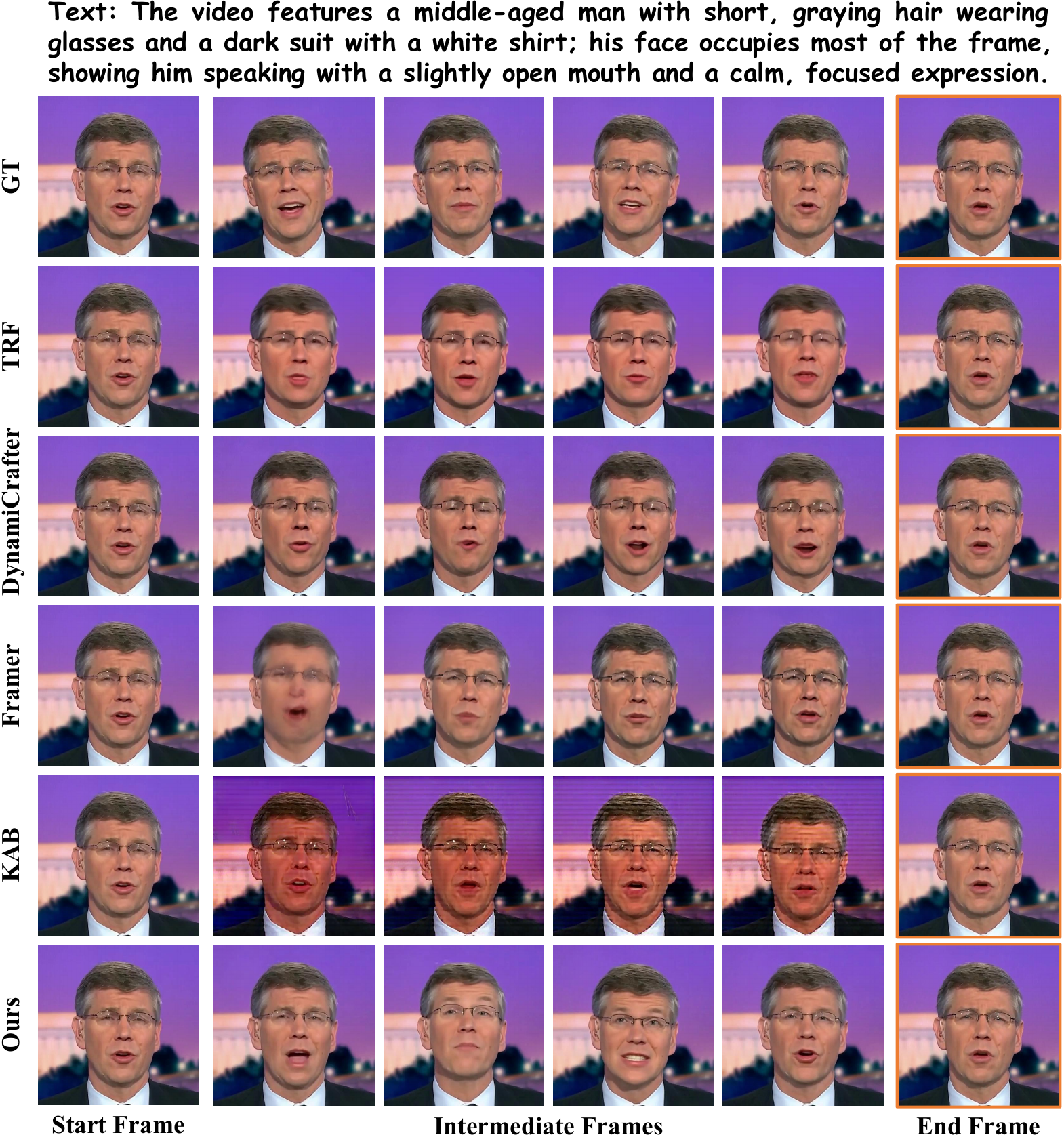}
    \vspace{-6mm}
    \caption{Qualitative comparison with video inbetweening models.  BBF produces richer and more natural facial expressions while maintaining coherent temporal transitions.
    }
    \label{fig:qual_hdtf_talking}
    \vspace{-4mm}
\end{figure}

\paragraph{Comparison with Talking-Head Generation Methods.}

\noindent
\textbf{\emph{Quantitative comparison}}.
We compare BBF with representative talking-head generation methods, including the GAN-based SadTalker~\cite{zhang2023sadtalker}, diffusion-based AniPortrait~\cite{wei2024aniportrait} and EchoMimic~\cite{chen2025echomimic}, SVD-based Sonic~\cite{ji2025sonic}, and Wan-based OmniAvatar~\cite{gan2025omniavatar} and StableAvatar~\cite{tu2025stableavatar}. Experiments are conducted on the HDTF and Hallo3 datasets, and the results are summarized in Table~\ref{tab:hdtf-hallo3}.

Most existing talking-head generation methods are designed for speech-driven animation from a single reference image and therefore cannot directly leverage both the start and end frames required for talking-head inbetweening. To enable a fair comparison, we additionally fine-tune Wan2.1 on our training set using the same start--end frame conditioning as BBF, denoted as FT-Wan2.1.

We randomly sample and crop 20 video clips (3--5 seconds each) from each dataset for evaluation. As shown in Table~\ref{tab:hdtf-hallo3}, BBF achieves the best performance on both datasets across most evaluation metrics. Compared with the strongest baseline, BBF reduces the FID and FVD on Hallo3 by 23.3\% and 36.5\%, respectively. Notably, BBF outperforms both existing talking-head generation models and the fine-tuned FT-Wan2.1, demonstrating that its superior performance stems not only from utilizing both boundary frames, but also from effectively integrating multimodal contextual cues to model realistic motion evolution.

\noindent
\textbf{\emph{Qualitative comparison}}.
Fig.~\ref{fig:Qualitative evaluation result} presents qualitative comparisons with representative talking-head generation methods. 
Existing talking-head generation methods (e.g., Sonic, OmniAvatar, and StableAvatar) often fail to produce intermediate frames that are consistent with the target ending frame, resulting in noticeable endpoint inconsistencies and abrupt motion transitions. 
Although FT-Wan2.1 is conditioned on both boundary frames, it is not explicitly designed to model motion evolution between them, leading to unstable intermediate structures and visual artifacts.
In contrast, by explicitly modeling motion evolution from the surrounding video context, BBF generates stable and realistic transitions while preserving fine-grained speech-driven facial dynamics.

\paragraph{Comparison with Video Inbetweening Methods.} 
\textbf{\emph{Comparison on Talking-Head Videos}}.
We further compare BBF with representative video inbetweening methods on the talking-head benchmark HDTF. 
As shown in Table~\ref{tab:davis_hdtf_results_with_latency}, BBF achieves the best overall performance across most evaluation metrics. 
Qualitative comparisons in Fig.~\ref{fig:qual_hdtf_talking} further demonstrate that BBF generates more realistic facial dynamics and temporally coherent transitions. 
These results suggest that generic video inbetweening methods, despite preserving global temporal consistency, are often insufficient for modeling the realistic facial dynamics required for talking-head videos. In contrast, by explicitly modeling motion evolution between the boundary frames, BBF produces more natural and temporally coherent intermediate frames.

\noindent
\textbf{\emph{Comparison on Generic Videos}}.
We further evaluate BBF on the DAVIS benchmark to assess its generalization to generic-scene video inbetweening.
As shown in Table~\ref{tab:davis_hdtf_results_with_latency}, BBF achieves the best overall performance across most evaluation metrics. 
Although FT-Wan2.1 obtains a marginally lower LPIPS, BBF consistently delivers higher visual fidelity and temporal coherence, as evidenced by its superior FID, FVD, PSNR, and SSIM scores. 
These results suggest that the proposed method is not limited to talking-head videos and can effectively generalize to generic video inbetweening scenarios.
Qualitative comparisons are provided in the Appendix.

\begin{figure}[t]
  \centering
  \vspace{-6pt}
  \includegraphics[width=\linewidth]{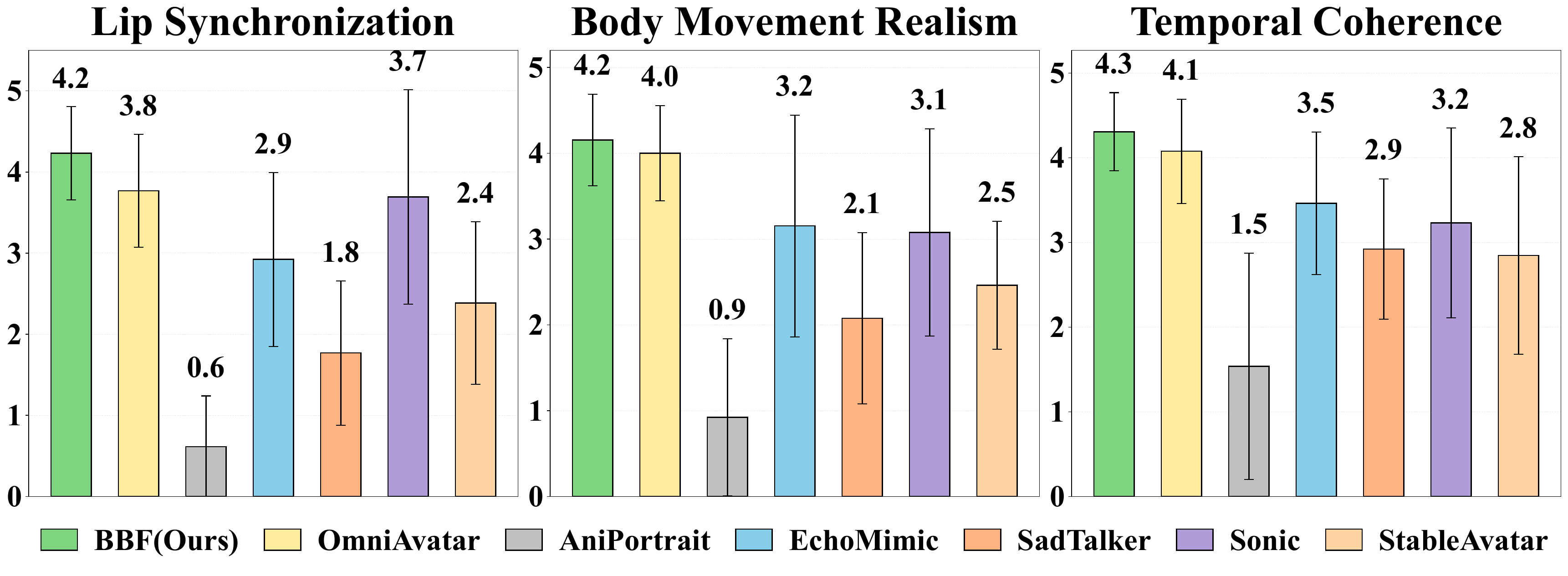}
  \vspace{-6mm}
  \caption{Human evaluation results across SOTA models.}
  \label{fig:Human_evaluation_result}
  \vspace{-4pt}
\end{figure}

\begin{table}[t]
\centering
\resizebox{\linewidth}{!}{%
\begin{tabular}{l c c c c c c}
  \toprule
  Cond. & FID$\downarrow$ & FVD$\downarrow$ & LPIPS$\downarrow$ & PSNR$\uparrow$ & SSIM$\uparrow$ & Sync-D$\downarrow$ \\
  \midrule
  Video clips & 29.53 & 780.12 & 0.18 & 21.75 & \textbf{0.75} & 13.63 \\
  Text        & 28.88 & 601.10 & 0.17 & 21.46 & 0.74 & 13.70 \\
  Audio       & 30.20 & 342.19 & \textbf{0.16} & 21.77 & \textbf{0.75} & 12.70 \\
  V \& T \& A & \textbf{26.42} & \textbf{335.67} & \textbf{0.16} & \textbf{22.41} & 0.73 & \textbf{12.59} \\
  \bottomrule
\end{tabular}%
}
\vspace{-2mm}
\caption{Effect of different input modalities on HDTF.}
\label{tab:hdtf_modalities}
\vspace{-4mm}
\end{table}

\begin{figure}[t]
    \centering
    \includegraphics[width=\columnwidth]
    {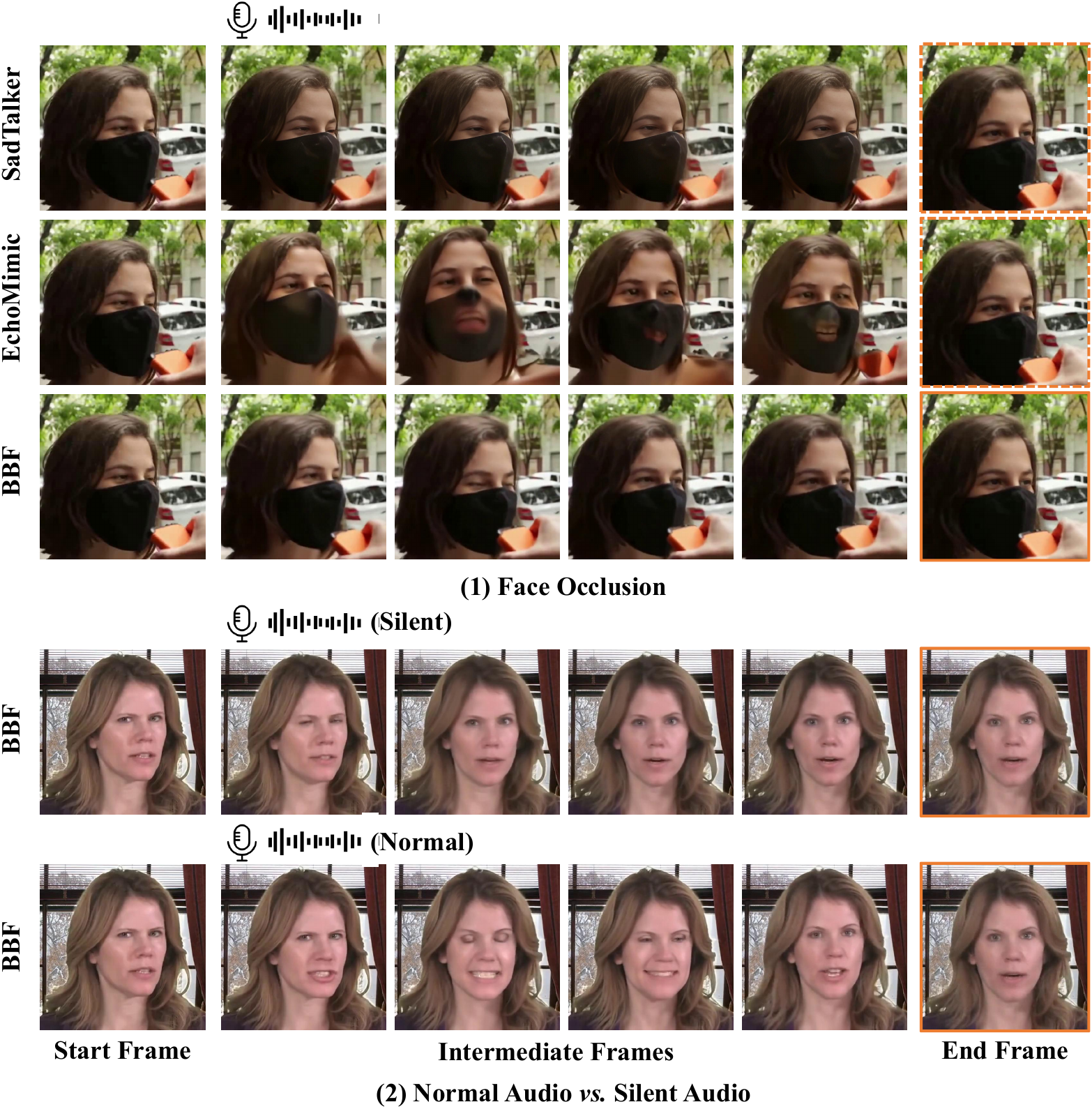}
    \vspace{-6mm}
    \caption{Robustness to degraded multimodal inputs. (1) BBF remains robust under severe face occlusion. (2) BBF suppresses unnecessary lip movements under silent audio while maintaining smooth facial dynamics.
    }
    \label{fig:qual_talking}
\vspace{-3mm}
\end{figure}

\paragraph{Human Evaluation.}

We conduct a human evaluation to compare BBF with SOTA methods. Thirty participants rated the generated videos using a 5-point Likert scale along three dimensions: lip synchronization, body movement realism, and temporal coherence. 
As shown in Fig.~\ref{fig:Human_evaluation_result}, BBF achieves the highest scores across all three dimensions, demonstrating its superior perceptual quality and temporal consistency, which aligns well with human subjective judgment.

\begin{table}[h]
\centering
\resizebox{\linewidth}{!}{%
\begin{tabular}{l c c c c c c}
  \toprule
  Masking Prob. & FID$\downarrow$ & FVD$\downarrow$ & LPIPS$\downarrow$ & PSNR$\uparrow$ & SSIM$\uparrow$ & Sync-D$\downarrow$ \\
  \midrule
  30\% & 10.74 & \textbf{174.24} & 0.15 & 22.11 & \textbf{0.78} & 13.39 \\
  10\% & 12.45 & 246.10 & 0.16 & 21.76 & 0.77 & 13.04 \\
  30\%$\rightarrow$10\% & \textbf{10.50} & 187.02 & \textbf{0.14} & \textbf{22.41} & \textbf{0.78} & \textbf{12.66} \\
  \bottomrule
\end{tabular}%
}
\vspace{-2mm}
\caption{Effect of audio masking schedules on HDTF.}
\label{tab:hdtf_paradigm}
\vspace{-4mm}
\end{table}

\subsection{Ablation Study}
\paragraph{Input Condition.}
To evaluate the contribution of each conditioning modality, we compare BBF using video-only, text-only, audio-only, and full multimodal conditioning. As shown in Table~\ref{tab:hdtf_modalities}, different modalities exhibit complementary strengths: text conditioning achieves the best fidelity (FID 28.88), while audio conditioning provides the best temporal alignment (Sync-D 12.70). Video-only conditioning performs competitively but remains inferior to multimodal conditioning overall. 
By jointly leveraging all modalities, BBF achieves the best overall performance, demonstrating that complementary multimodal cues are essential for generating realistic and temporally coherent talking-head videos.

\paragraph{Optimization Strategy.}
To evaluate the progressive optimization strategy, we compare three audio masking schedules on HDTF under 2,000 steps: constant 30\% masking, constant 10\% masking, and staged masking (30\% for the first 1,000 steps followed by 10\% for the remaining 1,000 steps). 
As shown in Table~\ref{tab:hdtf_paradigm}, the staged strategy achieves the best overall performance, demonstrating that the effectiveness of the proposed optimization strategy in balancing structural consistency and fine-grained motion learning.

\paragraph{Robustness to Degraded Modalities.}
To evaluate the robustness of BBF under degraded multimodal inputs, we provide qualitative results in Fig.~\ref{fig:qual_talking}. Under severe face occlusion (Fig.~\ref{fig:qual_talking} (1)), existing methods are easily misled, even hallucinating mouth movements on the mask and producing unrealistic facial structures. In contrast, BBF preserves plausible facial structures and smooth temporal transitions. When the speech signal is replaced with silent audio (Fig.~\ref{fig:qual_talking} (2)), BBF naturally suppresses unnecessary lip movements while maintaining coherent facial dynamics. These results demonstrate the robustness of BBF under both degraded visual and audio conditions.

\section{Conclusion}
In this paper, we present BBF, a talking-head inbetweening framework that generates realistic transitions between two existing video clips under multimodal conditioning. By jointly modeling endpoint constraints, motion evolution, and speech dynamics through decoupled multimodal conditioning, together with a progressive training strategy, BBF produces temporally coherent and visually realistic intermediate frames. Extensive experiments on both talking-head and generic video inbetweening benchmarks demonstrate consistent improvements over state-of-the-art methods.

\bibliography{BBF_arxiv}

\newpage

\appendix
\section{Appendix}
\label{sec:appendix}

\begin{table*}[t]
\centering
\begin{tabular}{l ccccc ccccc c}
  \toprule
  \multirow{2}{*}{\textbf{Model}} &
    \multicolumn{5}{c}{\textbf{HDTF}} &
    \multicolumn{5}{c}{\textbf{DAVIS}} &
    \multirow{2}{*}{\textbf{Lat.(s)$\downarrow$}} \\
  \cmidrule(lr){2-6}\cmidrule(lr){7-11}
  & FID$\downarrow$ & FVD$\downarrow$ & LPIPS$\downarrow$ & PSNR$\uparrow$ & SSIM$\uparrow$
  & FID$\downarrow$ & FVD$\downarrow$ & LPIPS$\downarrow$ & PSNR$\uparrow$ & SSIM$\uparrow$ & \\
  \midrule

  AMT           
                & 10.07 & 954.95 & \textbf{0.12} & 20.70 & 0.70 & 261.51 & 2443.07 & 0.45 & 13.79 & 0.44 & 29.96 \\
  VFIMamba      
                & \underline{9.91} & \underline{752.59} & \textbf{0.12} & \underline{21.72} & \textbf{0.78} & 236.65 & 1464.24 & 0.37 & 15.82 & \textbf{0.55} & 17.23 \\
  LBBDM         
                & 11.07 & 919.46 & \textbf{0.12} & 20.89 & 0.71 & 257.87 & 2437.30 & 0.41 & 13.44 & 0.44 & 405.16 \\
  EDEN          
                & \textbf{9.23} & 859.73 & \textbf{0.12} & 21.06 & 0.73 & \underline{227.97} & \underline{1259.42} & \underline{0.33} & \textbf{16.49} & \underline{0.54} & 19.33 \\
  \midrule
  BBF (Ours)
                 & 11.67 & \textbf{174.69} & 0.14 & \textbf{23.04} & \textbf{0.78} & \textbf{147.10} & \textbf{713.01} & \textbf{0.30} & \underline{15.38} & \underline{0.54} & 201.71 \\
  \bottomrule
\end{tabular}
\caption{Comparison with video frame interpolation models on the talking-head dataset HDTF and the generic-scene dataset DAVIS.}
\label{tab:davis_hdtf_VFI}
\end{table*}

\subsection{Qualitative Comparison on Generic Videos}
\begin{figure}[t]
  \centering
  \includegraphics[width=\columnwidth]{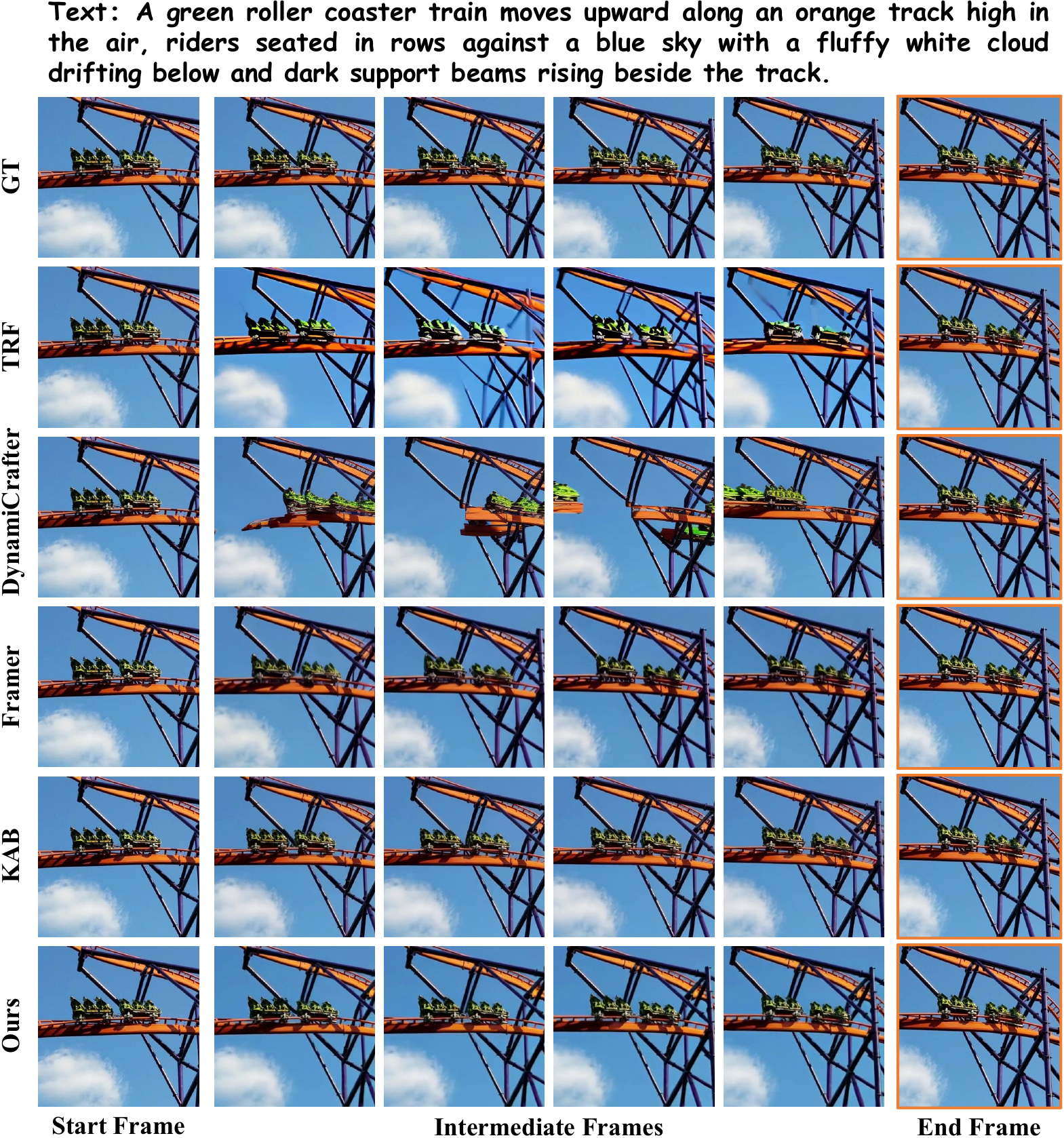}
  \vspace{-1mm}
  \caption{Qualitative comparison on the generic-scene DAVIS benchmark. Existing methods exhibit structural distortions, inconsistent object locations, or conservative motion progression. In contrast, BBF generates smoother and more realistic motion transitions. 
  \textbf{Some temporal differences are difficult to convey in static images and are more clearly demonstrated in the supplementary video.}}
  \vspace{-2mm}
  \label{fig:qual_generic}
\end{figure}

Fig. \ref{fig:qual_generic} presents qualitative comparisons with representative video inbetweening methods on the generic-scene DAVIS benchmark. 
In the roller-coaster sequence, TRF and DynamiCrafter produce noticeable structural distortions and inconsistent vehicle locations, while Framer and KAB exhibit relatively conservative motion progression despite preserving the overall scene geometry. 
In contrast, BBF generates smooth and realistic motion transitions while maintaining structural consistency. 
The temporal differences are more clearly demonstrated in the supplementary video. These results further demonstrate the strong generalization ability of BBF on generic video inbetweening.

\subsection{Comparison with Video Interpolation Methods}

We compare BBF with representative video frame interpolation (VFI) methods, including AMT~\cite{li2023amt}, VFIMamba~\cite{zhang2024vfimamba}, LBBDM, and EDEN~\cite{zhang2025eden}. 
To evaluate their performance in the long-range talking-head inbetweening setting, we sample 81-frame sequences from HDTF and 27-frame sequences from DAVIS, retain the first and last frames as boundary conditions, and remove the remaining intermediate frames. Since these methods are designed for interpolation between temporally adjacent frames, we follow Framer~\cite{wang2024framer} and apply them recursively to bridge the large temporal gap. All methods are evaluated on the same clips at a resolution of $512\times512$. Latency is measured as the end-to-end inference time for generating an 81-frame video on a single NVIDIA A100 GPU.

\paragraph{Quantitative Comparison.}
As shown in Table~\ref{tab:davis_hdtf_VFI}, BBF achieves the strongest overall performance among the evaluated video interpolation methods under this challenging long-range setting. On the talking-head dataset HDTF, BBF achieves the lowest FVD (174.69), the highest PSNR (23.04), and a tied-best SSIM (0.78), demonstrating superior temporal consistency and reconstruction quality. On the generic-scene dataset DAVIS, BBF further achieves the best FID (147.10), FVD (713.01), and LPIPS (0.30), while remaining competitive in PSNR and SSIM. These results demonstrate that BBF effectively models long-range motion evolution and generalizes well to generic video inbetweening scenarios.

\begin{figure}[t]
    \centering
    \includegraphics[width=\columnwidth]
    {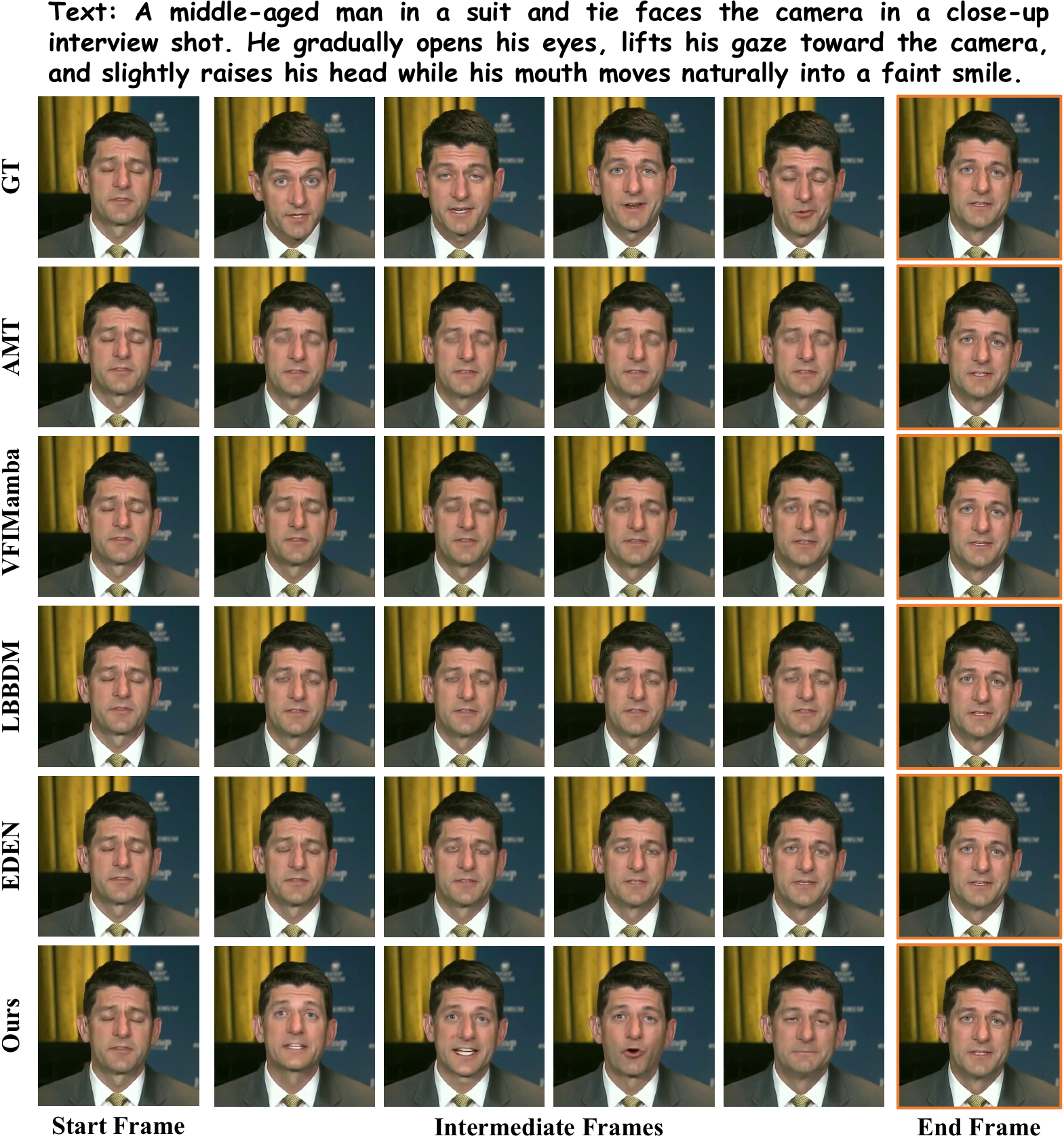}
    \vspace{-1mm}
    \caption{
    Qualitative comparison with video interpolation methods on the talking-head dataset HDTF. BBF recovers richer and more realistic facial dynamics.
    }
    \label{fig:qual_VFI_hdtf}
    \vspace{-2mm}
\end{figure}

\begin{figure}[t]
    \centering
    \includegraphics[width=\columnwidth]
    {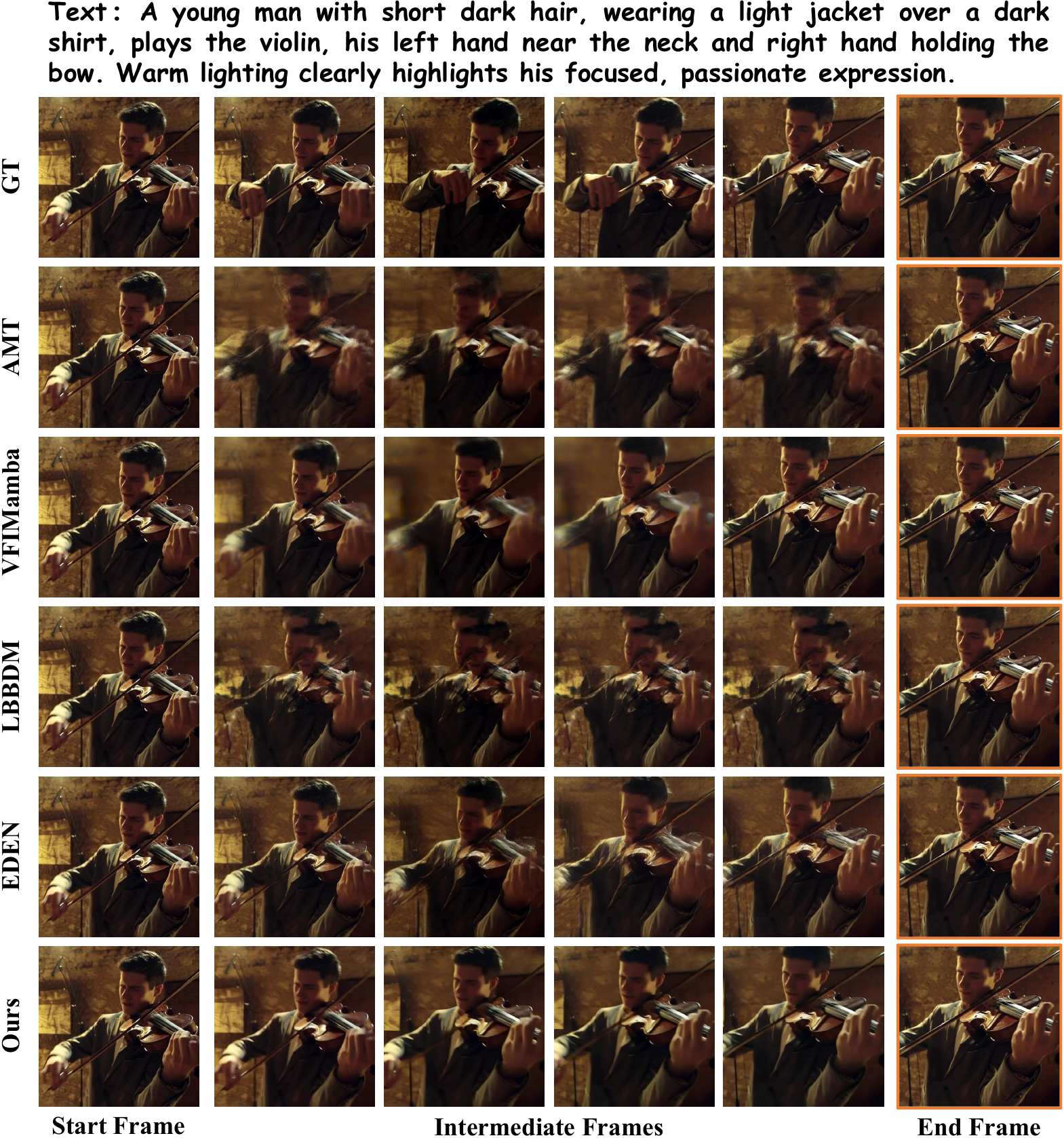}
    \vspace{-1mm}
    \caption{Qualitative comparison with video interpolation methods on the generic-scene dataset DAVIS. BBF better preserves object structures and generates more coherent motion trajectories under long-range motion.}
    \label{fig:qual_VFI_DAVIS}
    \vspace{-2mm}
\end{figure}

\paragraph{Qualitative Comparison.}
Figs.~\ref{fig:qual_VFI_hdtf} and~\ref{fig:qual_VFI_DAVIS} present qualitative comparisons on HDTF and DAVIS, respectively. On HDTF, the VFI baselines often produce over-smoothed transitions between the boundary states, suppressing intermediate variations in eye state, gaze, and facial expression. In contrast, BBF recovers richer facial dynamics while preserving identity and structural consistency. On DAVIS, the coordinated motion of the performer and instrument causes the VFI baselines to exhibit varying degrees of motion blur, ghosting, and structural distortion. BBF better preserves the structures of the subject, violin, and bow while generating a coherent motion trajectory. These observations complement the quantitative gains in FVD and further support the effectiveness of BBF for long-range video inbetweening.

\subsection{Limitations}

\paragraph{Computational Efficiency.}
As reported in Tables~\ref{tab:hdtf-hallo3} and \ref{tab:davis_hdtf_results_with_latency} of the main paper, BBF achieves inference latency comparable to existing diffusion-based talking-head generation and generative inbetweening methods. However, its diffusion-based video generation process inevitably incurs a relatively high computational cost, making the current implementation more suitable for offline editing than latency-sensitive applications. Future work will investigate few-step distillation, efficient sampling solvers, and temporal feature reuse to further improve inference efficiency and enable real-time deployment.

\paragraph{Generalization Scope.}
Although BBF performs competitively on both face-centric and generic benchmarks, its generalization beyond the current evaluation setting remains to be systematically investigated. Challenging scenarios such as multilingual speech, highly expressive facial motions, extreme viewpoints, and substantially longer temporal gaps may require more robust multimodal modeling. Future work will evaluate BBF on more diverse datasets and settings to further improve its generalization ability.

\end{document}